\def\year{2021}\relax
\documentclass[letterpaper]{article} % DO NOT CHANGE THIS
\usepackage{aaai21}  % DO NOT CHANGE THIS
\usepackage{times}  % DO NOT CHANGE THIS
\usepackage{helvet} % DO NOT CHANGE THIS
\usepackage{courier}  % DO NOT CHANGE THIS
\usepackage[hyphens]{url}  % DO NOT CHANGE THIS
\usepackage{graphicx} % DO NOT CHANGE THIS
\usepackage{natbib}  % DO NOT CHANGE THIS AND DO NOT ADD ANY OPTIONS TO IT
\usepackage{caption} % DO NOT CHANGE THIS AND DO NOT ADD ANY OPTIONS TO IT
\usepackage{array}
\usepackage{makecell}
\usepackage{multirow}
\usepackage{booktabs}
\usepackage{amsmath, amssymb}
\usepackage{multicol, blindtext}
\usepackage{caption}
\usepackage{multirow}
\usepackage{siunitx}
\usepackage{etoolbox}
\usepackage[labelsep=period]{caption}
\usepackage{siunitx}
\usepackage{subfigure}
\title{DenserNet: Weakly Supervised Visual Localization\\ Using Multi-Scale Feature Aggregation}
\author{
    Dongfang Liu\textsuperscript{\rm1}, Yiming Cui\textsuperscript{\rm2}\thanks{ is the corresponding author}, Liqi Yan\textsuperscript{\rm 3}, Christos Mousas\textsuperscript{\rm 1}, Baijian Yang\textsuperscript{\rm 1}, Yingjie Chen\textsuperscript{\rm 1}\\
}
\affiliations{
    \textsuperscript{\rm 1}{\textnormal{Purdue University}}\\ 
    \textsuperscript{\rm 2}{\textnormal{University of Florida}}\\
    \textsuperscript{\rm 3}{\textnormal{Fudan University}}\\
   liu2538@purdue.edu, cuiyiming@ufl.edu, yanliqi@westlake.edu.cn,\\ cmousas@purdue.edu, byang@purdue.edu, victorchen@purdue.edu
}
\begin{document}

\maketitle

\begin{abstract}
 In this work, we introduce a Denser Feature Network (DenserNet) for visual localization. Our work provides three principal contributions. First, we develop a convolutional neural network (CNN) architecture which aggregates feature maps at different semantic levels for image representations. Using denser feature maps, our method can produce more keypoint features and increase image retrieval accuracy. Second, our model is trained end-to-end without pixel-level annotation other than positive and negative GPS-tagged image pairs. We use a weakly supervised triplet ranking loss to learn discriminative features and encourage keypoint feature repeatability for image representation. Finally, our method is computationally efficient as our architecture has shared features and parameters during forwarding propagation. Our method is flexible and can be crafted on a light-weighted backbone architecture to achieve appealing efficiency with a  small penalty on accuracy. Extensive experiment results indicate that our method sets a new state-of-the-art on four challenging large-scale localization benchmarks and three image retrieval benchmarks with the same level of supervision. The code is available at \url{https://github.com/goodproj13/DenserNet}.
\end{abstract}
\section{Introduction}
The task of visual localization is to predict the geographic location of a query image, based on its comparisons to GPS-tagged images from a database \cite{salarian2018improved}. Visual localization has drawn considerable attention recently due to its potential value to wide-ranging applications such as robot navigation or autonomous driving \cite{liu2020visual, liu2020video}. Under the region where GPS signal is partially or completely shadowed, visual localization is an effective addition to GPS to support the operation of these mobile agents.\\
\begin{figure}[tb]
  \centering
\subfigure[]{
\includegraphics[width=3.8cm]{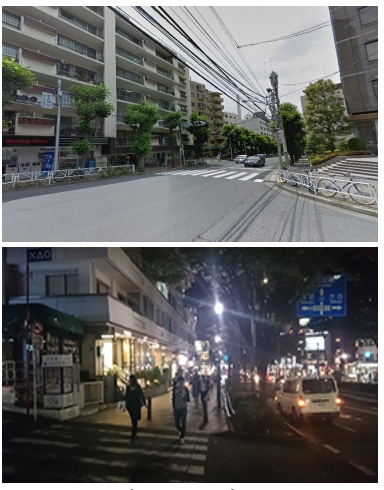}
}
\subfigure[]{
\includegraphics[width=3.88cm]{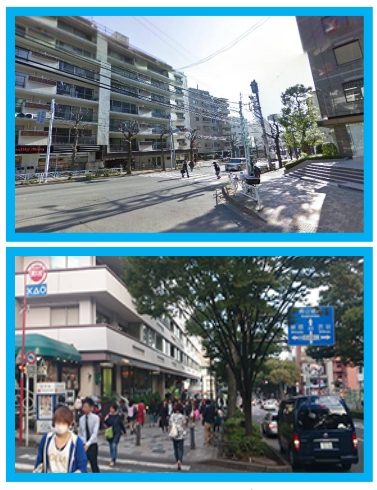}
}
  \caption{DenserNet can correctly perform visual localization under challenging conditions. Despite environmental variations (such as pedestrian or vehicle occlusions, illumination changes, and seasonal changes), DenserNet can find the location (a) from the database based on the query (b). 
%   \textcolor{red}{Since you first mention (b) and then (a), why don't you swap the images so you mention first the (a) and then the (b). It is not a major issue, but more of an aesthetic correction}
  }
  \label{Figure 1}
\end{figure}
\indent In our work, we cast the visual localization problem as an image retrieval task \cite{arandjelovic2016netvlad}.  Image retrieval task relies on local features to search over a GPS-tagged image database to estimate the current location.  The primary challenge is how to produce discriminative image representation so that images from nearby locations would have similar representations in feature spaces while images from different locations would have dissimilar representations. Typically, a large-scale database contains images of similar man-made structures or landmarks that may cause severe ambiguities. Illumination variations or occlusions may also change object appearances which compromise the localization prediction \cite{ tolias2016image}.  To address these problems, our method is designed to be robust for a large-scale dataset with different challenging conditions (Figure \ref{Figure 1}).\\
\indent In the last decade, convolutional neural networks (CNNs) have emerged as a powerful technique to explore image representations including visual localization \cite{sandler2018mobilenetv2, simonyan2014very}. CNN-based visual localization networks have a similar architecture \cite{dusmanu2019d2, jin2017learned}. They generally have a convolutional backbone encoder to produce feature maps of the input image. Then, a detection-description decoder organizes the obtained feature maps to depict the image representation. The conventional approaches only use the feature maps from a single semantic level and fail to exploit multi-scale features from different semantic levels. This limitation motivates our approach to exploit features from multi-level semantics to improve localization behavior.\\
\begin{figure*}[tb]
  \centering
\includegraphics[width=15cm, trim={0 0.3cm 0 0.4cm},clip]{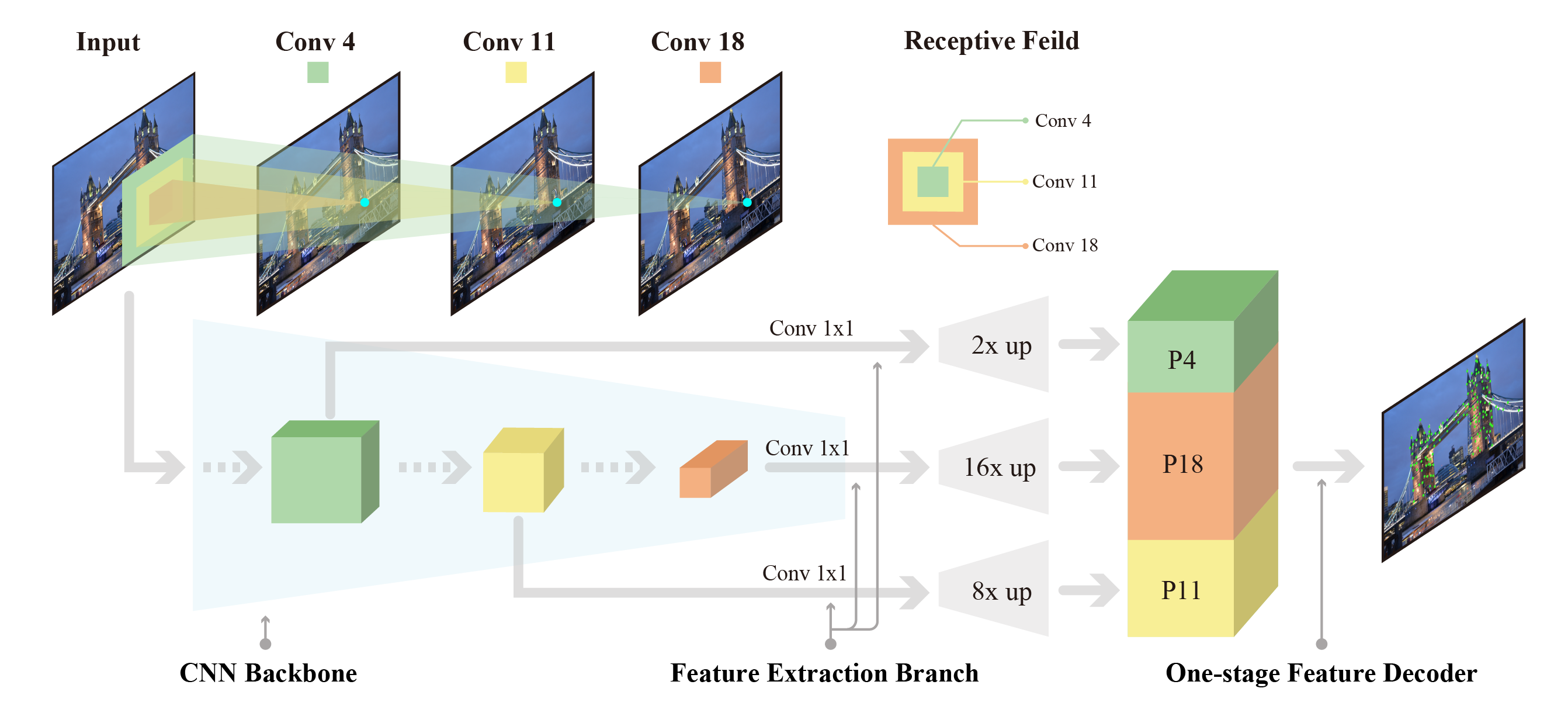}
  \vskip -6pt
  \caption{The demonstration of DenserNet using MobileNet backbone \cite{sandler2018mobilenetv2}. The proposed method can aggregate multi-level feature maps and thus produce significantly more local keypoint features than concurrent methods \cite{arandjelovic2016netvlad, dusmanu2019d2}. For simplicity,
$3 \times 3$ convolution layers after upsampling are not shown.
  }
  \label{Figure 2}
  \vskip -12pt
\end{figure*}
\indent So far, many state-of-the-art methods are trained using pixel-level annotations to supervise the process of feature learning \cite{luo2020aslfeat,dusmanu2019d2}. However, the ground truth correspondences between image pairs are expensive to obtain at the scale required to train a CNN-based method. Also, such supervised priors based on human annotations may not fully capture all relevant features for image representation to train a network \cite{jin2017learned}. Thus, training a CNN-based method with strong supervision is not efficient and effective enough for generalization. In contrast, an increasing number of studies \cite{liu2019stochastic, jin2017learned} employ a weakly supervised manner for training using image-level labels. They compute the feature distance between the query-positive and the query-negative image pairs to discover imagery discriminativeness. These prior explorations impact our work.\\
\indent Building on the lessons learned from the concurrent approaches \cite{arandjelovic2016netvlad, detone2018superpoint}, 
% we want to investigate whether we can aggregate feature maps from different semantic levels to produce dense local features for better image representation. This research question requires us to address the following three main challenges. First, how to design a CNN architecture so we can access multi-level feature maps from the different locations of the backbone network for learning image representation? Second, how to effectively aggregate dense features while remaining within the computational constraints? Third, how to effectively train the developed network tailored for the visual localization task in an end-to-end manner? To address these challenges, 
our work brings the following three contributions. First, we introduce a Denser Feature Network (DenserNet), a novel CNN-based method for visual localization tasks. We find inspiration from DenseNet \cite{huang2017densely} to aggregate feature maps at different semantic levels 
% \textcolor{red}{what are these semantic levels? Can you explain?} 
for image representations, as shown in Figure \ref{Figure 2}. 
% The architecture of DenserNet follows the paradigm of conventional visual localization methods such as NetVLAD\cite{arandjelovic2016netvlad} and CRN\cite{jin2017learned}, but which has a CNN-based encoder for feature extraction and a decoder to produce image representation. 
Our method is an intuitive extension of  NetVLAD \cite{arandjelovic2016netvlad} by adding feature extraction branches to obtain multi-scale features, in parallel with the existing backbone network. The feature extraction branches aggregate features from the lower-level, the mid-level, and the higher-level layer of the backbone network. Compared to conventional methods \cite{dusmanu2019d2, jin2017learned, detone2018superpoint}, our approach is able to produce denser keypoint features which increase the number of inlier matches between image pairs and, in turn, improve the matching accuracy under challenging conditions. Extensive experiment results indicate that our method improves the visual localization performance on several benchmarks by a large margin.\\
\indent Second, we propose a weakly supervised approach for training. We design a modified triplet ranking loss to effectively organize obtained features and predict image representations. Our training requires no expensive pixel-wise ground truths other than GPS-tagged images, which are easy to obtain. In our design, task-relevant keypoint features are discovered in an unsupervised fashion, as the proposed method is able to learn in which context should be suppressed or emphasized to achieve better location recognition. Inspired by \cite{dusmanu2019d2}, our training method performs a joint optimization for both detection and description tasks, which encourages the repeatability of discriminative detection and improves the description accuracy.\\
\indent Finally, DenserNet is computationally efficient. Since the three feature extraction branches are all based on the same CNN backbone, they have shared features and parameters during computation. Theoretical and empirical evaluations demonstrate that, although DenserNet uses extra sub-network branches to aggregate more feature maps for the location inference, it only requires a limited additional computation load. Thus, our method can achieve efficient inference and remain within the computational constraints.
\section{Related Work 
% \textcolor{red}{This section looks so limited. In most papers the RW section is almost a page. Can you expand? Maybe move parts from your Introduction to RW section.}
}
\indent Recent advances in CNNs have made it possible to use local keypoint features of an image to predict location hypothesis \cite{arandjelovic2016netvlad}. Although the CNN-based approaches are proved to be effective for visual localization, they struggle to perform well when the feature homogeneousness occurs substantially \cite{liu2019stochastic}. Modern man-made structures frequently have architectural similarities so it is difficult for the conventional methods to handle this challenging situation \cite{salarian2018improved}. In addition, the performance of conventional methods typically degrades under extreme environmental changes, such as illumination or landmark scale changes \cite{dusmanu2019d2}.\\
\indent A critical reason causing these problems for conventional methods \cite{dusmanu2019d2} is that they only use features from one semantic level for prediction. Thus, they fail to exploit features from different levels of semantics to capture more multi-scale details. This semantic gap introduces a critical problem in both feature learning and predicting \cite{sarlin2019coarse}. To address the above challenges, our method is designed to produce more keypoint features using feature extraction branches to organically extract features from different semantic levels. Compared to the latest similar work~\cite{luo2020aslfeat} that uses strong supervision for training and has a heavy working pipeline for inference, our method is simple yet effective. We can achieve competitive performance using a weak supervision fashion without any bells and whistles. 
% Since our feature extraction branches are paralleled to the backbone network, the proposed method is computationally efficient for features sharing. We aggregate features with both strong semantics and high spatial resolution to obtain discriminative representation.  Thus our method is robust to environmental changes such as illumination and landmark scale.
\section{DenserNet}
\subsection{Overview}
The architecture of DenserNet is shown in Figure \ref{Figure 2}. DenserNet includes a CNN backbone, three feature extraction branches, and a one-stage feature decoder. Our design leverages the intrinsic nature of modern CNN architecture which can produce rich hierarchical features from different convolutional layers from a single forward pass. Thus we can obtain multi-scale features with low additional costs to close the semantic gap in feature learning.
% Inspired by \cite{sarlin2019coarse}\cite{huang2017densely}, our approach has a backbone encoder with three feature extraction branches. 
Each feature extraction branch sticks out at different layers of the backbone network to extract features from different semantic levels. We aggregate the obtained features to capture strong image representation. Intuitively, our method is more robust to scale variances than methods learned from single-scale features and thus improves the localization performance. We elaborate on the detail of our method below.
% the input image. While producing more useful keypoint features, our network leverages shared feature of the  backbone network thus requires limited additional computation.
\subsection{Feature Extraction Branch}
The design of DenserNet is flexible and allows various backbone options. For fast runtime, we use MobileNetV2 \cite{sandler2018mobilenetv2}, a light-weighted network, as our backbone network. DenserNet has the lower-level, the mid-level, and the higher-level branch based on different locations of the backbone. Specifically, the lower-level and mid-level branches stick out at the \texttt{conv4} 
% \textcolor{red}{I would use the texttt{} for all convs in this paper (see the conv4).}
and \texttt{conv11} respectively, and the higher-level branch is attached to the \texttt{conv18}.\\
\indent The input features for each feature extraction branch are $\{C4, C11, C18\}$, having strides of $\{4, 16, 32\}$ pixels with respect to the input image. Each feature extraction branch employs a modified SuperPoint layer \cite{detone2018superpoint}. Instead of having the keypoints and local descriptors from the original implementation, the modified SuperPoint only increases the channel dimensions and upsamples the feature maps. The SuperPoint layer has a shallow structure of a non-linear $1\times1$ convolutional layer and a upsampling layer. The non-linear $1\times1$ convolutional layer with \texttt{ReLU6} 
% \textcolor{red}{also for ReLu I would use the texttt} 
activation is used to increase the channel dimensions of each branch. The upsampling layer increases the spatial resolution of the feature maps in a non-learned manner, which is much faster than using transposed convolutions \cite{sarlin2019coarse}. After SuperPoint layer, the output feature maps are $\{P4, P11, P18\}$ corresponding to $\{C4, C11, C18\}$. Compared to $\{C4, C11, C18\}$, the channel dimensions for $\{P4, P11, P18\}$ increase 3, 1.5, 1.1 times respectively, and their spatial resolutions is brought back to half of the input resolution, which retains more landmark details. Following the practice from \cite{lin2017feature}, we also append a $3 \times 3$ convolution on each output in order to reduce the aliasing effect from upsampling. The obtained features from each feature extraction branch are aggregated by channel-wise concatenation. The aggregated features 
have channel dimensions of 520. Afterwards, the dense features are fed into the one-stage feature decoder to produce image representation,
% \textcolor{red}{Can you explain a bit here. I think i get lost.}
 which is discussed in the next section.\\
\indent The core concept for our design is to leverage the dense features to enhance image representations. Following the simple rule, our approach can use many design choices. Adding more feature extraction branches (more than three) can improve the prediction performance but demand more memory footprint. Empirically, our method has the best runtime and accuracy tradeoff. Our design is flexible to any CNN backbone. To achieve higher accuracy, we also implement a VGG16 \cite{simonyan2014very} version of DenserNet.
\begin{figure}[tb]
  \centering
\includegraphics[width=8cm]{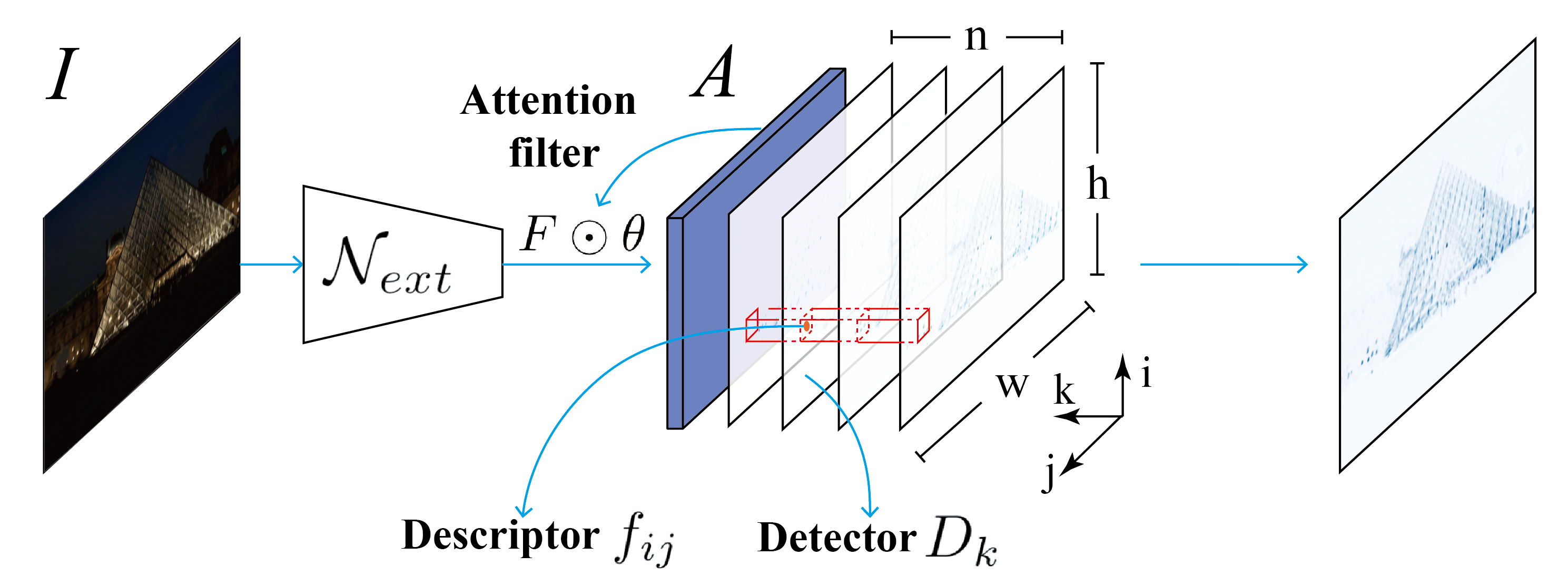}
  \caption{One-stage feature decoder. We use the attention filter to produce attention features and couple descriptor and detector together to delineate image representation.}
  \label{Figure 3}
\end{figure}
\subsection{One-stage Feature Decoder}
\indent The feature extraction branches and the backbone network  work together as the feature extraction encoder ${\mathcal{N}}_{ext}$ that processes the input image {${I}$} and produces the dense feature maps ${F} = {\mathcal{N}}_{ext} \left({I}\right)$, ${F} \in {\mathbb{R}}^{h\times w \times n}$ ($h\times \textit{w}$ is the feature map size and $n$ is the number of channels). Similarly to \cite{dusmanu2019d2}, we employ a one-stage feature decoder to delineate the image representation. Our one-stage feature decoder includes the feature attention filter, the feature descriptor, and the feature detector as shown in Figure \ref{Figure 3}.\\
% performs detection and description simultaneously to prevent information loss in the quantization step.\\
\indent \textbf{Feature attention filter.} 
Feature attention filter implicitly encodes spatial concept into the parameters of the attention features, which are flexible to represent the irregular landmarks. The attention features are computed by matrix multiplication of dense features $F$ and attention filters $\theta$:
\begin{equation}
{
    A =\mathbb{R}(F\odot\theta),A \in \mathbb{R}^{h\times w \times n}, 
    }
    \label{Equation 0}
\end{equation}
where $A$ is the attention features and $\mathbb{R}$ is the \texttt{ReLU} activation. Based on CAM~\cite{zhou2016learning}, the attention filter $\theta$ 
screens out the spatial locations with pedestrians, vehicles, and vegetation on the feature maps.
% indicate which spatial locations should be focused for feature detection and description.
\\
\indent \textbf{Feature descriptor.} The attention feature maps $A$ can be expressed by a set of descriptor vectors $f$:
\begin{equation}
{
    f_{ij} =A_{ij:},f_{ij} \in \mathbb{R}^{n}, 
    }
    \label{Equation 1}
\end{equation}
where $i \in \mathbb{R}^{h}$ and $j \in \mathbb{R}^{w}$. These descriptor vectors can be used to establish feature correspondences by calculating the Euclidean distance between images. 
% We L2-normalize the obtained descriptors to be unit length.
We use $L2$ normalization to make the obtained descriptors in unit length.
Then, we properly adjust these descriptors in training so that the same points for a scene produce similar descriptors,  which robustly describe discriminative appearance variations.\\
\indent \textbf{Feature detector.} In the same vein, the attention feature maps $A$ can be expressed as a collection of detector $D$:
\begin{equation}
{
    D_k = {A}_{::k}, \quad {D}_{k} \in {\mathbb{R}}^{h\times w}
    }
    \label{Equation 2}
\end{equation}
where $k \in \mathbb{R}^{n}$. In this expression, the feature extraction network $\mathcal{N}_{ext}$ produces $n$ different detection response maps $D_k$. If pixel point $(i, j)$ is detected, we denote a hard feature detector $D_{(ij)k}$ which is the most strong detection in all channels. We then perform a channel-wise softmax around its neighbours to obtain the local softmax score:
\begin{equation}
{s}_{ij}= \frac{{\exp(D}_{(ij)k})}{\sum_{i'=i-1}^{i+1}{\sum_{j'=j-1}^{j+1}{ \exp(D}_{(i'j')k})}}
.\end{equation}
Finally, we compute an image-level normalization for the softmax score to obtain the detection score at a pixel $(i,j)$:
\begin{equation}
    % s_{ij} = \frac{s_{ij}}
    % {\sum_{(i',j')\in (h,w)} s_{i'j'}}
     \widetilde{s}_{ij} = \frac{{s}_{ij}}{\sum_{i'=1}^{h}{\sum_{j'=1}^{w}{ s_{i'j'}}}}
    \label{Equation 3}
.\end{equation}
\subsection{Time Complexity Analysis}
Similar to the conventional methods \cite{arandjelovic2016netvlad, detone2018superpoint},  our method has a backbone encoder for feature extraction and feature decoder for image representation. However, our approach also includes three feature extraction branches stick out from the lower, middle, higher level of the backbone network to efficiently produce denser features. Unlike patch-based networks MNV \cite{sarlin2018leveraging} and LF-Net \cite{ono2018lf}  who adopt a Siamese sub-network to produce more features with a high computational cost, our approach has shared features across the three branches, which effectively avoids the  computation overhead.\\
\indent For runtime complexity, the ratio of our approach versus the conventional methods is:
\begin{equation}
    r = \frac{ \mathcal{O}\big(\mathcal{N}_{cnn}\big)+3\times\mathcal{O}\big(\mathcal{N}_{br}\big)+\mathcal{O}\big(\mathcal{N}_{de}\big)}
    {\mathcal{O}\big(\mathcal{N}_{cnn}\big)+\mathcal{O}\big(\mathcal{N}_{de}\big)}.
    \label{Equation 5}
\end{equation}
where $\mathcal{O}\big(\mathcal{N}_{cnn}\big)$, $\mathcal{O}\big(\mathcal{N}_{br}\big)$, and $\mathcal{O}\big(\mathcal{N}_{de}\big)$ are the function complexity for the backbone encoder, the feature extraction branch, and the feature decoder for image representation. Since  $\mathcal{O}\big(\mathcal{N}_{de}\big) \ll \mathcal{O}\big(\mathcal{N}_{cnn}\big)$, the ratio can be approximated as: $r \approx 1+ \frac{3\times\mathcal{O}\big(\mathcal{N}_{br}\big)}
    {\mathcal{O}\big(\mathcal{N}_{cnn}\big)}.$
Compared to the conventional methods, the increased computational cost of the proposed method mainly comes from $\mathcal{N}_{br}$ branches. This increased computational cost can be ignored because $\mathcal{O}\big(\mathcal{N}_{br}\big) \ll \mathcal{O}\big(\mathcal{N}_{cnn}\big)$. Thus, our approach is appealing for its efficiency. We report empirical results in the later runtime evaluation section to resonate with our analysis here.
\begin{figure}[tb]
  \centering
\includegraphics[width=8.5cm]{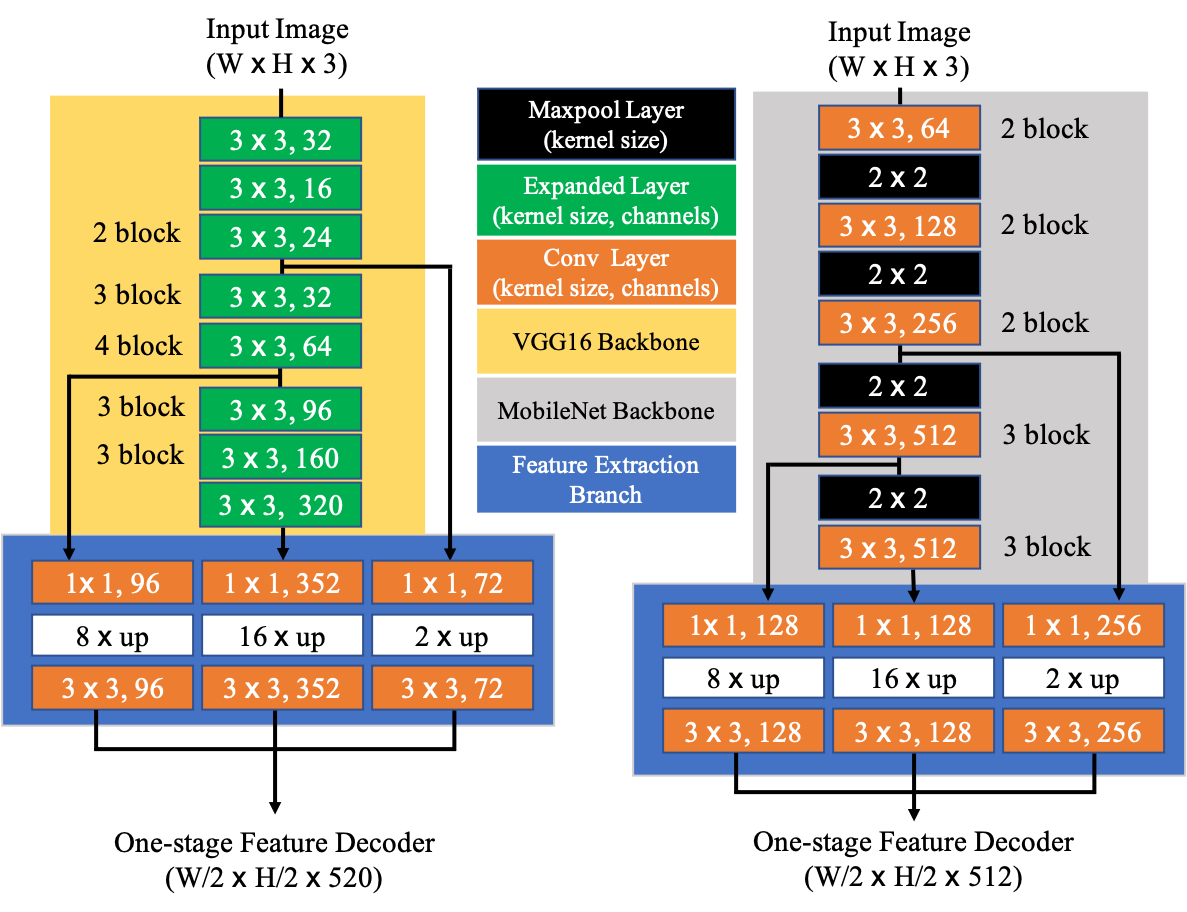}
  \caption{Architectures of the proposed method.}
  \label{Arch}
\end{figure}
\subsection{Model Architecture}
\indent In our experiments, we implement two models of variants which are based on  MobileNetV2~\cite{sandler2018mobilenetv2} and VGG-16~\cite{simonyan2014very}. The two architectures are displayed in Figure~\ref{Arch}. The MobileNet-based method is discussed in the section of Feature Extraction Branch. For VGG-based method, its feature extraction branches stick out at Con3\_2, Con4\_3, and Con5\_3 respectively, and their corresponding feature maps are~$\{P3\_2,P4\_3,P5\_3\}$. In each feature extraction branch, the channel dimensions for $\{P3\_2,P4\_3,P5\_3\}$ are multiplied 1,  0.25,  0.25  times respectively, and their spatial resolutions are brought back to one-fourth of the input resolution. The obtained features from each feature extraction branch are aggregated by channel-wise concatenation. 
% \begin{figure}[tb]
%   \centering
% \includegraphics[width=8.5cm]{LaTeX/img/arc.png}
%   \caption{Architectures of the proposed method.}
%   \label{Arch}
% \end{figure}
\subsection{Training Objective}
We propose a modified triplet ranking loss to jointly optimize our detectors and descriptors in a weakly supervised manner, which only requires cheap image triplets for training. In our setting, the location of a query image $I_q$ is approximated by searching for the nearest neighbors in feature space among the reference images $\{I_r\}$. Thus, the objective for training our method is to match the positive references $\{I^+_r\}$ which are closer to the query image and vice-versa to the negative ones $\{I^-_r\}$. \\
\indent For a pair of images  $({I}^{1}, {I}^{2})$ and potential corresponding feature points
$P:p_1 \leftrightarrow p_2$ between them (where $p_1 \in {I}^{1}, p_2 \in {I}^{2}$), we want to minimize
the distance $\sum_{p \in \mathcal{P}}\parallel {f}_{p}^{1} - {f}_{p}^{2}\parallel_{2}$ of the corresponding descriptors between the positive pairs while maximize the distance between the negative ones. In order to increase the repeatability of detection \cite{dusmanu2019d2}, we also add a detection term to encourage repeatability of effective detection between two images:\\
\begin{equation}
    \mathcal{R}\left(I^{1}, I^{2}\right) =  \sum_{p \in P}\frac{\widetilde{s}^{1}_{p} \widetilde{s}^{2}_{p}}{\sum_{p' \in P}\widetilde{s}^{1}_{p'}\widetilde{s}^{2}_{p'}} \parallel {f}_{p}^{1} - {f}_{p}^{2} \parallel_{2}.
    \label{Equation 6}
\end{equation}
where $P$ is the set of all corresponding feature points between ${I}^{1}$ and ${I}^{2}$. $\widetilde{s}^{1}_{p}$ and $\widetilde{s}^{2}_{p}$ are the detection scores in (4) at each corresponding  feature  point of the paired images. Accordingly, our training objective can be defined as:\\
\begin{equation}
\mathcal{L_{R}}\left(I_t, I^+_r, I^-_r\right)=\max\left(M + \mathcal{R}\left(I_{t}, I^+_{r}\right) - \mathcal{R}\left(I_{t}, I^-_{r}\right),0\right)
\label{Equation 7}.
\end{equation}
where $I_{t}$, $I^+_{r}$, and $I^-_{r}$ are the training query image, the positive reference, and the negative reference respectively. In order to minimize the proposed loss, the distances of the discriminative descriptors between the training query and the positive reference are encouraged to be small and the associated detection scores are enforced to be large. Using the weakly supervised fashion for training, our method effectively learns feature representation pertaining to which features should be suppressed or emphasized.

\section{Experiment and Results}
\indent In this section, we first describe the implementation details and the evaluation datasets. We then use an ablation study to investigate the improvements of our method from the baseline method. Next, we test the proposed method on several tasks in comparison with some of the state-of-the-art methods. Finally, we conclude with the runtime evaluation.
\subsection{Implementation Details}
\textbf{Training data mining.} We train the proposed method by using Pitts30k-training dataset \cite{arandjelovic2016netvlad}. 
Following \cite{arandjelovic2016netvlad, jin2017learned}, we group the positive $\{I^+_r\}$ and negative $\{I^-_r\}$ images for each training query image $I_t$. The positive images are the closest neighbors to each query image in the feature space at its nearby geo-locations, while the negative image is far away. Our training data mining purposefully selects positive images with fewer dynamic objects (i.e. pedestrians or bicycles). We will demonstrate that the stationary training data is beneficial to improve the localization behavior in the experiment. The 30K training query images generate four image triplets. Thus, we obtain a total of 120K image triplets with 112K for training and 8K for validation.\\
\indent\textbf{Training process.}
All experiments are performed on a workstation with an Intel Core i7-7820X CPU and four NVIDIA GeForce GTX 3080Ti GPU. Both VGG16 and MobileNetV2 based methods are pretrained on ImageNet \cite{deng2009imagenet}. In training, we exploit standard data augmentation in training, such as motion blur, random Gaussian noise, brightness changes to improve the robustness of our  methods to illumination variations and viewpoint changes. Specifically, the  margin  $M$ is set at 0.1, 30 epochs  are performed  using batch size of 4 triplets, Adam \cite{kingma2014adam} with the learning rates of $10^{-3}$ which is halved every 6 epochs, momentum of 0.9, and weight decay of $10^{-3}$. We use the Precision-Recall curve to evaluate the training performance \cite{arandjelovic2016netvlad}. A query is considered to be correct if at least one result from the top $N$ retrieved database images is within d = 25 meters from the ground truth position of the query image. We use the best method (the highest $recall@5$) on the validation for testing.
\begin{figure*}[tb]
  \centering
\includegraphics[width=18cm]{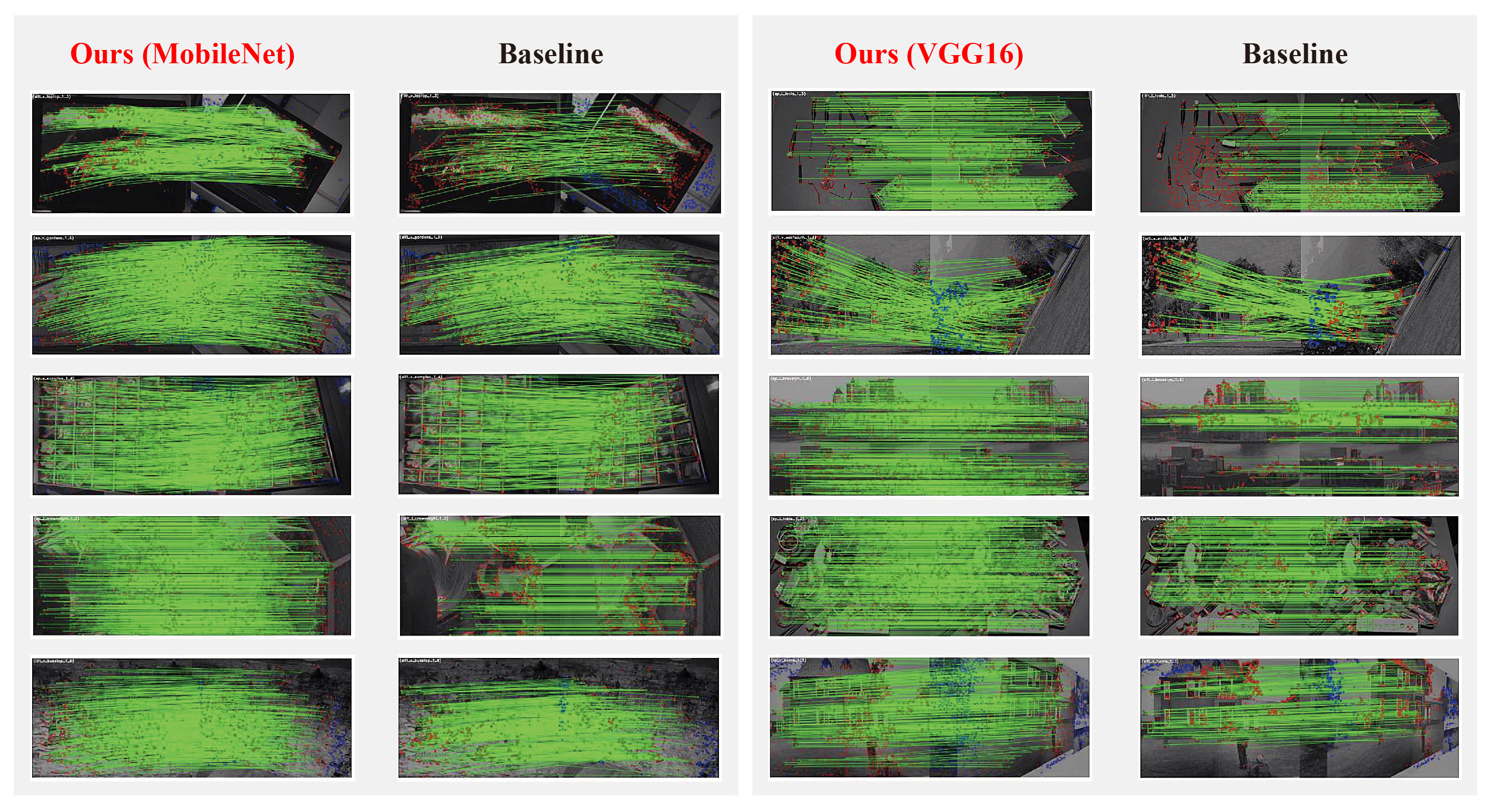}
  \caption{Qualitative examples on the HPatches dataset. The green lines indicate all correct correspondences which are repeatable. DenserNet can produce denser and more correct matches compared to the baseline method. Red dots are mis-matched points and blue dots are not visible to the shared viewpoint region of the image pairs.  Best viewed in the digital format with zoom.}
  \label{Figure 4}
\end{figure*}
\subsection{Evaluation Datasets and Metrics}
We assess our method on three different tasks with a number of publicly available benchmarks.\\
\indent \textbf{Feature matching task.} HPatches dataset \cite{balntas2017hpatches} is adopted to evaluate the effectiveness of feature extraction and matching. Following \cite{detone2018superpoint}, we use the detection repeatability and mean localization error (MLE) of the keypoint features to evaluate the detector; and we use the mean average precision (mAP) and the matching score (MS) to evaluate the descriptor. The mAP assesses a method's ability to suppress spurious matches. MS reflects the overall performance of both the detector and the descriptor.\\
\indent \textbf{Large-scale visual localization task.} Pitts250k-test \cite{torii2013visual}, Tokyo 24/7 \cite{torii201524}, TokyoTM-val \cite{arandjelovic2016netvlad}, and and Sf-0 \cite{chen2011city} are used to investigate the performance of visual localization. For the first three benchmarks, a query is considered to be correct if at least one result from the top $N$ retrieved database images is within d = 25 meters from the ground truth position of the query image. For Sf-0, the query result is considered to be correct  if at least one result from the top $N$ retrieved database images has the same building ID.\\ 
\indent \textbf{Image retrieval task.} Since we cast the visual localization problem as an image retrieval task, we also evaluate our method on image retrieval benchmarks. Oxford 5k \cite{philbin2007object}, Paris 6k \cite{philbin2008lost}, and Holidays \cite{jegou2008hamming} is used to test the generalization of our method for image representations on image retrieval. We use mAP for evaluation.
\begin{table}[!tb]
    % \scriptsize
    \newcolumntype{P}[1]{>{\centering\arraybackslash}p{#1}}
    \footnotesize
    \centering
    \begin{tabular}{c|c|c|c|c|c|c|c}
    \toprule
         \multirow{2}{*}{Meth.} &  \multirow{2}{*}{HB} & \multirow{2}{*}{LB} & \multirow{2}{*}{MB} & \multicolumn{2}{c|}{Det. Metric} & \multicolumn{2}{c}{Des. Metric} \\
      \cline{5-8}
      \rule{0pt}{10pt} &  &  & & Rep. & MLE & mAP & MS\\
      \midrule
      a~(M) & $\surd$ & & & 0.556 & 0.99 & 0.782 & 0.448\\
      b~(M) & $\surd$ & $\surd$ & & 0.568 & 1.02 & 0.799 & 0.456 \\
      c~(M) & $\surd$ & & $\surd$ & 0.573 & 1.03 & 0.808 & 0.461 \\
      d~(M) & $\surd$ & $\surd$ & $\surd$ & 0.592 & 1.04 & 0.827 & 0.473 \\
      \midrule
      a~(V) & $\surd$ & & & 0.578 & 1.12 & 0.822 & 0.462\\
      b~(V) & $\surd$ & $\surd$ & & 0.588 & 1.14 & 0.847 & 0.471 \\
      c~(V) & $\surd$ & & $\surd$ & 0.596 & 1.15 & 0.859 & 0.475 \\
      d~(V) & $\surd$ & $\surd$ & $\surd$ & 0.628 & 1.17 & 0.886 & 0.481 \\
      \bottomrule
    \end{tabular}
    \caption{Evaluation results from HPatches dataset. We use HPatches to examine the ability for feature extraction and matching. Method a is the baseline while method b and c are its variants. Method d is the proposed DenserNet. The results indicate the improvements of DenserNet from baseline in the all metrics. M and V in methods stand for MobileNet and VGG-based backbones respectively. HB, LB,  MB stand for higher-level, lower-level, and mid-level branch respectively.}
    \label{tab:Table 1}
\end{table}

\subsection{Ablation Study}
\begin{figure}[tb]
  \centering
\includegraphics[width=8cm, trim={1.2cm 0 1.5cm 0},clip]{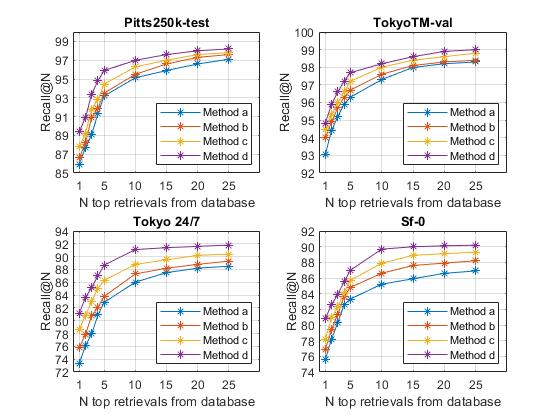}
  \caption{Ablation results of Recalls at $N$ top retrievals with all VGG-based methods.}
  \label{Figure 5}
\end{figure}

\begin{figure}[tb]
  \centering
\includegraphics[width=8cm, trim={1.1cm 0 1.5cm 0},clip]{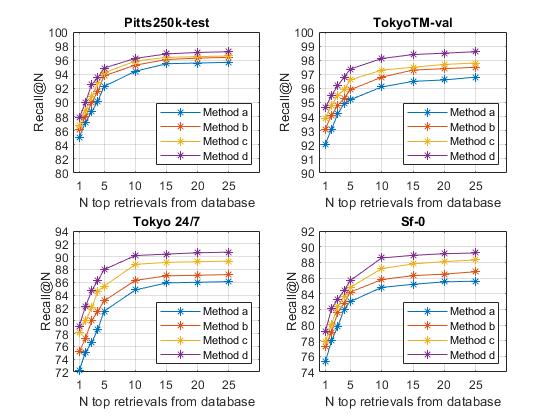}
  \caption{ Ablation results of recalls at $N$ top retrievals with all MobileNet-based methods.}
  \label{MBM}
\end{figure}

We conduct an ablation study to validate the improvements of DenserNet from a strong baseline and its variants. Method (a) is the baseline which only has the higher-level branch for feature extraction. It is identical to SuperPoint \cite{detone2018superpoint}. Method (b) and (c) are two variants of the baseline method with the additional lower-level or mid-level branch respectively. Method (d) is the proposed method. The ablation study is conducted on feature matching and visual localization tasks.\\
\indent The results for feature matching are demonstrated in Table \ref{tab:Table 1}. As expected, VGG-based methods outperform their MobileNet-based counterparts because VGG16 has a better feature extraction capacity. From method (a) to method (d), adding the feature branch can give a boost in the performance of both detector and descriptor. The repeatability of keypoint features increases with the increase of the feature extraction branch while errors only increase with a small scale. Meanwhile, mAP and matching scores also increase accordingly by adding the feature extraction branch. Figure \ref{Figure 4} demonstrates the qualitative results of the VGG-based method. Our method can produce more keypoint features for effective matching than the baseline method.\\
\indent The results for all VGG-based methods on large-scale visual localization tasks are demonstrated in Figure \ref{Figure 5}. Method (d) outperforms the baseline method and its variants on all benchmarks. We observe a steady increase of $recall@N$ from method (a) to method (d). Based on the Precision-Recall curve, using an extra feature extraction branch is proved to be effective to improve the performance of localization as methods having more feature extraction branches achieve better results on all benchmarks. The results for all MobileNet-based methods on large-scale visual localization tasks are demonstrated in Figure \ref{MBM}.
\subsection{Comparison with State-of-the-Art Methods}
\begin{figure}[tb]
  \centering
\includegraphics[width=8cm, trim={1.1cm 0 1.5cm 0},clip]{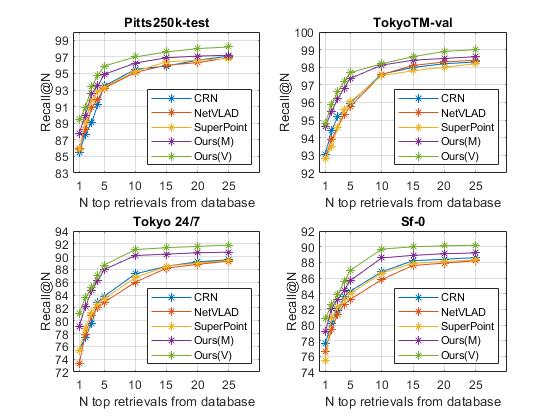}
  \caption{Comparison of recalls at $N$ top retrievals with state-of-the-art methods. Ours~(V) has VGG backbone and ours~(M) has MobileNet backbone.}
  \label{CSOT}
\end{figure}

To demonstrate the advancement of the proposed method, we compare our method with state-of-the-art methods on large-scale localization benchmarks and image retrieval tasks. We choose three leading methods NetVLAD, CRN, and SuperPoint for comparison. In order to have a fair comparison, we retrain all the methods with the same setup.
\begin{table*}[]
    \centering
    % \scriptsize
    \footnotesize
    \begin{tabular}{c|c|c|c|c|c|c|c|c|c|c|c|c}
    \toprule
    \multirow{2}{*}{Method} & \multicolumn{3}{c|}{Pitts 250k-test} & \multicolumn{3}{c|}{TokyoTM-val} & \multicolumn{3}{c|}{Tokyo 24/7} &
    \multicolumn{3}{c}{Sf-0}\\
    \cline{2-13}
    \rule{0pt}{10pt} & r@1 & r@5 & r@10 & r@1 & r@5 & r@10 & r@1 & r@5 & r@10 & r@1 & r@5 & r@10\\
    \midrule
    Ours~(VGG) & 89.40 & 95.90 & 96.99 & 94.80 & 97.69 & 98.20 & 81.20 & 88.67 & 91.10 & 80.80 & 86.99 & 89.68 \\
    Ours~(MobileNet) & 87.82 & 94.89 & 96.26 & 94.62 & 97.37 & 98.12 & 79.17 & 87.99 & 90.17 & 79.12 & 85.68 & 88.56 \\
    CRN & 85.50 & 93.49 & 95.50 & 93.07 & 95.97 & 97.61 & 75.39 & 83.81 & 87.31 & 77.62 & 84.31 & 86.80\\
    NetVLAD & 85.95 & 93.21 & 95.13 & 92.85 & 95.77 & 97.59 & 73.33 & 82.86 & 86.03 & 76.57 & 83.27 & 85.80\\
    SuperPoint & 85.78 & 93.36 & 95.26 & 92.81 & 96.04 & 97.53 & 75.37 & 83.44 & 86.73 & 75.52 & 84.01 & 86.60\\
     \bottomrule
    \end{tabular}
    \caption{Comparison of Recalls at $N$ top retrievals of different methods on the four large-scale visual localization benchmarks. 
    % We highlight the \textcolor{red}{best} and \textcolor{blue}{second-best} methods.
    }
    \label{tab:Table 2}
\end{table*}
% \subsubsection{Large-scale Localization Benchmark Results} 
\indent \textbf{Large-scale localization benchmark results.} We report the $recall@N$ results for different methods in Table \ref{tab:Table 2}. Both our VGG16 and MobileNet-based methods consistently outperform state-of-the-art methods by a significant margin on all benchmarks. For instance, on the Pitts250k dataset, our improvements over the next best method NetVLAD at $r@1$ is 3.45\% for the VGG-based architecture and 1.87\% for the MobileNet-based architecture. On the Sf-0 dataset, our VGG16 and MobileNet-based methods achieve $r@1$ of 80.08\% and 79.12\% respectively compared to the next best method CRN by a margin of 3.18\% and 1.5\% respectively. Our methods also obtain similar improvements on Tokyo 24/7 and Tokyo TM. The results from all benchmarks confirm our assumption for this proposed work: leveraging denser features from multiple semantic levels and using the right supervision for training, the proposed method can effectively learn discriminative yet compact image representations for visual localization. More detailed comparisons are displayed in Figure \ref{CSOT}.
\begin{figure}[!tb]
  \centering
\includegraphics[width=8.5cm]{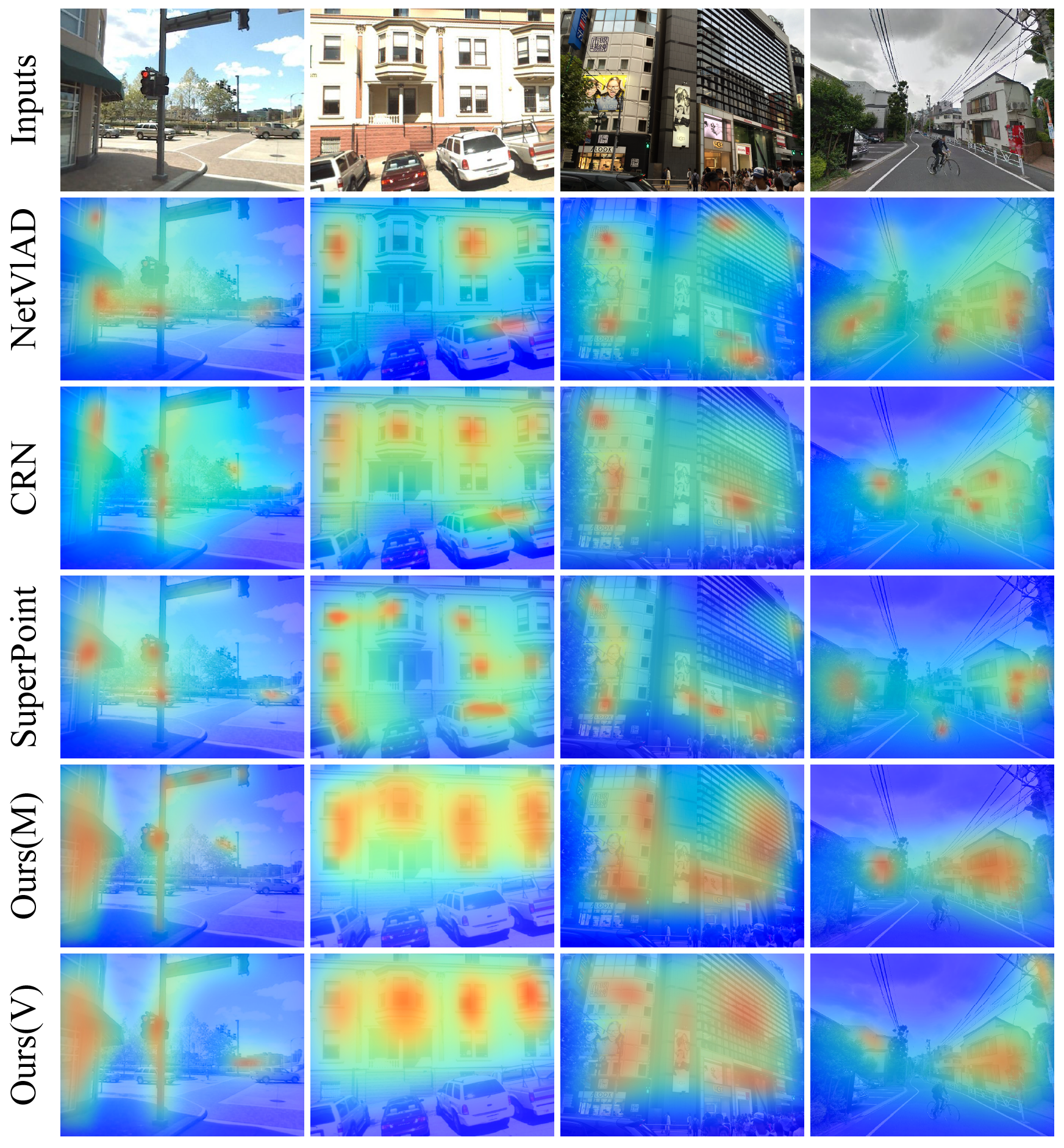}
  \caption{Comparison of feature emphasis.  With the feature attention filter and training data mining, our methods focus on the distinctive details of buildings, while avoiding confusing visual  clues such  as  pedestrians,  vegetation,  or  vehicles  which  are  hard  for  feature repeatability.}
  \label{Figure 6}
\end{figure}
\begin{table}[tb]
\newcolumntype{P}[1]{>{\centering\arraybackslash}p{#1}}
    \centering
    % \scriptsize
    \footnotesize
    % \begin{tabular}{c|c|c|c|c|c|c}
    \begin{tabular}{P{1.3cm}|P{0.6cm}|P{0.6cm}|P{0.6cm}|P{0.6cm}|P{0.6cm}|P{0.6cm}}
    \toprule
    \multirow{2}{*}{Method}& \multicolumn{2}{c|}{Oxford 5k} & \multicolumn{2}{c|}{Paris 6k} & \multicolumn{2}{c}{Holidays} \\
    \cline{2-7}
    \rule{0pt}{10pt} & full & crop & full & crop & orig & rot\\
    \midrule
    Ours~(V) & 67.66 & 69.40 & 75.03 & 78.21 & 84.71 & 88.30 \\
    Ours~(M) & 66.88 & 68.26 & 75.00 & 78.30 & 84.01 & 88.28 \\
    CRN & 63.95 & 65.52 & 72.88 & 75.85 & 83.19 & 87.30\\
    NetVLAD & 63.09 & 65.33 & 72.53 & 75.67 & 82.67 & 86.83\\
    SuperPoint & 63.14 & 65.50 & 72.83 & 75.10 & 82.92 & 86.90\\
     \bottomrule
    \end{tabular}
    \caption{Comparison  with state-of-the-art methods for compact image representations (256-D) on image retrieval tasks. }
    \label{tab:Table 3}
\end{table}\\
\indent For qualitative analysis, we visualize the regions of the input image which are emphasized for localization prediction (as shown in Figure \ref{Figure 6}). Particularly, we use the heatmaps \cite{grun2016taxonomy} to highlight the feature emphasis of different regions on the input image. Based on the results, we observe that our method is superior to its counterparts in identifying more useful features for localization, despite the significant environmental variations in viewpoint or illumination. Our method focuses on the distinctive details of buildings that are identifiable for visual localization while avoiding confusing visual clues. In contrast, other methods generally focus on local features independently which are inherently limited. Many features are laid on confusing scenes such as pedestrians, vegetation, or vehicles which are hard for feature repeatability. We argue that our multi-scale feature aggregation, feature attention filter, training data mining, and training supervision technique collaboratively contribute to the improved localization behavior.\\
% The results indicate that our method scales well to visual localization under large-scale environments.\\
\indent \textbf{Image retrieval task results.} To assess the generalizability of our approach, we evaluate our methods trained only on Pitts30k \cite{arandjelovic2016netvlad} without any fine-tuning on the standard image retrieval datasets. For Oxford 5k \cite{philbin2007object} and Paris 6k \cite{philbin2008lost}, we use both the full and cropped images; for Holidays \cite{jegou2008hamming}, we use original and rotated images. The results are displayed in Table \ref{tab:Table 3}. Our results set the state-of-the-art for compact image representations (256-D) on all three datasets. On all metrics, our margin consistently over the mAP of other methods is 1-4\%. For example, there are a 3.71\% (our VGG-based method) and 2.93\% (our MobileNet-based method) improvements on Oxford 5k (full) than the next best method; there are a 3.88\% (our VGG-based method) and 2.74\% (our MobileNet-based method) improvements on Oxford 5k (crop) than the next best method. Since our methods only see building-oriented images, our results can be further improved by fine-tuning using the natural landmark images from the three image retrieval datasets.
\begin{table}[tb]
\newcolumntype{P}[1]{>{\centering\arraybackslash}p{#1}}
    \centering
    % \scriptsize
    \footnotesize
    \begin{tabular} {c|P{1.3cm}|c|c|c|P{0.5cm}|P{0.6cm}}
    \toprule
    \multirow{2}{*}{Dataset} & \multirow{2}{*}{Method} & \multicolumn{4}{c|}{Runtime(ms)} & \multirow{2}{*}{r@1}\\
    \cline{3-6}
    % & & \rotatebox{90}{Backbone encoder } & \rotatebox{90}{Feature  extraction branch } & \rotatebox{90}{Feature decoder }& \rotatebox{90}{Total} &\\
    \rule{0pt}{10pt} & & BE & FEB & FD &Total &\\
    \midrule
    % \multirow{5}{*}{Pitts250k-test} 
    \multirow{5}{*}{\makecell[c]{Pitts250k\\-test}} & Ours~(V) & 92 & 6 & 2 & 100 & 89.40 \\
    & Ours~(M) & 15 & 6 & 2 & 23 & 87.82\\
    & CRN & 92 & - & 6 & 98 & 85.50\\
    & NetVLAD & 92 & - & 5 & 97 & 85.95\\
    & SuperPoint & 43 & - & 7 & 50 & 85.78\\
    \midrule
    \multirow{5}{*}{Tokyo 24/7} & Ours~(V) & 96 & 7 & 3 & 106 & 81.20 \\
    & Ours~(M) & 18 & 7 & 3 & 28 & 79.17\\
    & CRN & 96 & - & 8 & 104 & 75.39\\
    & NetVLAD & 96 & - & 7 & 103 & 73.33\\
    & SuperPoint & 49 & - & 9 & 58 & 75.37\\
     \bottomrule
    \end{tabular}
    \caption{Runtime report. BE, FEB, and FD means backbone encoder, feature extraction branch, and feature decoder respectively.}
    \label{tab:Table 4}
\end{table}\\
\subsection{Runtime Evaluation} In this work, we provide two versions of our method. VGG-based model is designed to achieve high performance in accuracy while our MobileNet-based model has a better efficiency with a small penalty on accuracy. We analyze its runtime and compare it with other state-of-the-art methods. All the measurements are conducted on the same workstation. As shown in Table \ref{tab:Table 4}, the increase of our computation cost stems from the feature extraction branch (FEB). One advantage of our method is that our FEB shares features and avoids runtime overhead. The increased computational cost is affordable as it is much smaller than that of the backbone encoder and has little impact on the total runtime cost. Specifically, our MobileNet-based models are around 4$\times$ faster than CRN and NetVLAD which have the VGG-based architecture. For our VGG-based model, our speed is comparable to CRN and NetVLAD which have the same backbone network. Although our VGG-based models are slower than SuperPoint which uses a smaller backbone architecture, our methods outperform counterparts in localization $r@1$ by a large margin.
\section{Conclusion}
\indent In this work, we propose DenserNet, a novel CNN-based architecture that aggregates denser features from multiple semantics to achieve strong image representation. Results from extensive experiments indicate that our method is competitive with the current state-of-the-art methods on large-scale localization tasks with the same level of supervision. 
% \bibliographystyle{plain}
% \balance
% \section{Acknowledgment}
% We want to thank Dr. Laixi Shi from Carnegie Mellon University for her kind advice on our project and review of our manuscripts.
\bibliography{ref}

\begin{thebibliography}{28}
\providecommand{\natexlab}[1]{#1}
\providecommand{\url}[1]{\texttt{#1}}
\providecommand{\urlprefix}{URL }
\expandafter\ifx\csname urlstyle\endcsname\relax
  \providecommand{\doi}[1]{doi:\discretionary{}{}{}#1}\else
  \providecommand{\doi}{doi:\discretionary{}{}{}\begingroup
  \urlstyle{rm}\Url}\fi

\bibitem[{Arandjelovic et~al.(2016)Arandjelovic, Gronat, Torii, Pajdla, and
  Sivic}]{arandjelovic2016netvlad}
Arandjelovic, R.; Gronat, P.; Torii, A.; Pajdla, T.; and Sivic, J. 2016.
\newblock NetVLAD: CNN architecture for weakly supervised place recognition.
\newblock In \emph{Proceedings of the IEEE conference on computer vision and
  pattern recognition}, 5297--5307.

\bibitem[{Balntas et~al.(2017)Balntas, Lenc, Vedaldi, and
  Mikolajczyk}]{balntas2017hpatches}
Balntas, V.; Lenc, K.; Vedaldi, A.; and Mikolajczyk, K. 2017.
\newblock HPatches: A benchmark and evaluation of handcrafted and learned local
  descriptors.
\newblock In \emph{Proceedings of the IEEE Conference on Computer Vision and
  Pattern Recognition}, 5173--5182.

\bibitem[{Chen et~al.(2011)Chen, Baatz, K{\"o}ser, Tsai, Vedantham,
  Pylv{\"a}n{\"a}inen, Roimela, Chen, Bach, Pollefeys et~al.}]{chen2011city}
Chen, D.~M.; Baatz, G.; K{\"o}ser, K.; Tsai, S.~S.; Vedantham, R.;
  Pylv{\"a}n{\"a}inen, T.; Roimela, K.; Chen, X.; Bach, J.; Pollefeys, M.;
  et~al. 2011.
\newblock City-scale landmark identification on mobile devices.
\newblock In \emph{CVPR 2011}, 737--744. IEEE.

\bibitem[{Deng et~al.(2009)Deng, Dong, Socher, Li, Li, and
  Fei-Fei}]{deng2009imagenet}
Deng, J.; Dong, W.; Socher, R.; Li, L.-J.; Li, K.; and Fei-Fei, L. 2009.
\newblock Imagenet: A large-scale hierarchical image database.
\newblock In \emph{2009 IEEE conference on computer vision and pattern
  recognition}, 248--255. Ieee.

\bibitem[{DeTone, Malisiewicz, and Rabinovich(2018)}]{detone2018superpoint}
DeTone, D.; Malisiewicz, T.; and Rabinovich, A. 2018.
\newblock Superpoint: Self-supervised interest point detection and description.
\newblock In \emph{Proceedings of the IEEE Conference on Computer Vision and
  Pattern Recognition Workshops}, 224--236.

\bibitem[{Dusmanu et~al.(2019)Dusmanu, Rocco, Pajdla, Pollefeys, Sivic, Torii,
  and Sattler}]{dusmanu2019d2}
Dusmanu, M.; Rocco, I.; Pajdla, T.; Pollefeys, M.; Sivic, J.; Torii, A.; and
  Sattler, T. 2019.
\newblock D2-Net: A Trainable CNN for Joint Detection and Description of Local
  Features.
\newblock \emph{arXiv preprint arXiv:1905.03561} .

\bibitem[{Gr{\"u}n et~al.(2016)Gr{\"u}n, Rupprecht, Navab, and
  Tombari}]{grun2016taxonomy}
Gr{\"u}n, F.; Rupprecht, C.; Navab, N.; and Tombari, F. 2016.
\newblock A taxonomy and library for visualizing learned features in
  convolutional neural networks.
\newblock \emph{arXiv preprint arXiv:1606.07757} .

\bibitem[{Huang et~al.(2017)Huang, Liu, Van Der~Maaten, and
  Weinberger}]{huang2017densely}
Huang, G.; Liu, Z.; Van Der~Maaten, L.; and Weinberger, K.~Q. 2017.
\newblock Densely connected convolutional networks.
\newblock In \emph{Proceedings of the IEEE conference on computer vision and
  pattern recognition}, 4700--4708.

\bibitem[{Jegou, Douze, and Schmid(2008)}]{jegou2008hamming}
Jegou, H.; Douze, M.; and Schmid, C. 2008.
\newblock Hamming embedding and weak geometric consistency for large scale
  image search.
\newblock In \emph{European conference on computer vision}, 304--317. Springer.

\bibitem[{Jin~Kim, Dunn, and Frahm(2017)}]{jin2017learned}
Jin~Kim, H.; Dunn, E.; and Frahm, J.-M. 2017.
\newblock Learned contextual feature reweighting for image geo-localization.
\newblock In \emph{Proceedings of the IEEE Conference on Computer Vision and
  Pattern Recognition}, 2136--2145.

\bibitem[{Kingma and Ba(2014)}]{kingma2014adam}
Kingma, D.~P.; and Ba, J. 2014.
\newblock Adam: A method for stochastic optimization.
\newblock \emph{arXiv preprint arXiv:1412.6980} .

\bibitem[{Lin et~al.(2017)Lin, Doll{\'a}r, Girshick, He, Hariharan, and
  Belongie}]{lin2017feature}
Lin, T.-Y.; Doll{\'a}r, P.; Girshick, R.; He, K.; Hariharan, B.; and Belongie,
  S. 2017.
\newblock Feature pyramid networks for object detection.
\newblock In \emph{Proceedings of the IEEE conference on computer vision and
  pattern recognition}, 2117--2125.

\bibitem[{Liu et~al.(2020{\natexlab{a}})Liu, Cui, Chen, Zhang, and
  Fan}]{liu2020video}
Liu, D.; Cui, Y.; Chen, Y.; Zhang, J.; and Fan, B. 2020{\natexlab{a}}.
\newblock Video Object Detection For Autonomous Driving: Motion-aid Feature
  Calibration.
\newblock \emph{Neurocomputing} .

\bibitem[{Liu et~al.(2020{\natexlab{b}})Liu, Cui, Guo, Ding, Yang, and
  Chen}]{liu2020visual}
Liu, D.; Cui, Y.; Guo, X.; Ding, W.; Yang, B.; and Chen, Y. 2020{\natexlab{b}}.
\newblock Visual Localization for Autonomous Driving: Mapping the Accurate
  Location in the City Maze.

\bibitem[{Liu, Li, and Dai(2019)}]{liu2019stochastic}
Liu, L.; Li, H.; and Dai, Y. 2019.
\newblock Stochastic Attraction-Repulsion Embedding for Large Scale Image
  Localization.
\newblock In \emph{Proceedings of the IEEE International Conference on Computer
  Vision}, 2570--2579.

\bibitem[{Luo et~al.(2020)Luo, Zhou, Bai, Chen, Zhang, Yao, Li, Fang, and
  Quan}]{luo2020aslfeat}
Luo, Z.; Zhou, L.; Bai, X.; Chen, H.; Zhang, J.; Yao, Y.; Li, S.; Fang, T.; and
  Quan, L. 2020.
\newblock Aslfeat: Learning local features of accurate shape and localization.
\newblock In \emph{Proceedings of the IEEE/CVF Conference on Computer Vision
  and Pattern Recognition}, 6589--6598.

\bibitem[{Ono et~al.(2018)Ono, Trulls, Fua, and Yi}]{ono2018lf}
Ono, Y.; Trulls, E.; Fua, P.; and Yi, K.~M. 2018.
\newblock LF-Net: learning local features from images.
\newblock In \emph{Advances in neural information processing systems},
  6234--6244.

\bibitem[{Philbin et~al.(2007)Philbin, Chum, Isard, Sivic, and
  Zisserman}]{philbin2007object}
Philbin, J.; Chum, O.; Isard, M.; Sivic, J.; and Zisserman, A. 2007.
\newblock Object retrieval with large vocabularies and fast spatial matching.
\newblock In \emph{2007 IEEE conference on computer vision and pattern
  recognition}, 1--8. IEEE.

\bibitem[{Philbin et~al.(2008)Philbin, Chum, Isard, Sivic, and
  Zisserman}]{philbin2008lost}
Philbin, J.; Chum, O.; Isard, M.; Sivic, J.; and Zisserman, A. 2008.
\newblock Lost in quantization: Improving particular object retrieval in large
  scale image databases.
\newblock In \emph{2008 IEEE conference on computer vision and pattern
  recognition}, 1--8. IEEE.

\bibitem[{Salarian et~al.(2018)Salarian, Iliev, Cetin, and
  Ansari}]{salarian2018improved}
Salarian, M.; Iliev, N.; Cetin, A.~E.; and Ansari, R. 2018.
\newblock Improved Image-Based Localization Using SFM and Modified Coordinate
  System Transfer.
\newblock \emph{IEEE Transactions on Multimedia} 20(12): 3298--3310.

\bibitem[{Sandler et~al.(2018)Sandler, Howard, Zhu, Zhmoginov, and
  Chen}]{sandler2018mobilenetv2}
Sandler, M.; Howard, A.; Zhu, M.; Zhmoginov, A.; and Chen, L.-C. 2018.
\newblock Mobilenetv2: Inverted residuals and linear bottlenecks.
\newblock In \emph{Proceedings of the IEEE conference on computer vision and
  pattern recognition}, 4510--4520.

\bibitem[{Sarlin et~al.(2019)Sarlin, Cadena, Siegwart, and
  Dymczyk}]{sarlin2019coarse}
Sarlin, P.-E.; Cadena, C.; Siegwart, R.; and Dymczyk, M. 2019.
\newblock From coarse to fine: Robust hierarchical localization at large scale.
\newblock In \emph{Proceedings of the IEEE Conference on Computer Vision and
  Pattern Recognition}, 12716--12725.

\bibitem[{Sarlin et~al.(2018)Sarlin, Debraine, Dymczyk, Siegwart, and
  Cadena}]{sarlin2018leveraging}
Sarlin, P.-E.; Debraine, F.; Dymczyk, M.; Siegwart, R.; and Cadena, C. 2018.
\newblock Leveraging deep visual descriptors for hierarchical efficient
  localization.
\newblock \emph{arXiv preprint arXiv:1809.01019} .

\bibitem[{Simonyan and Zisserman(2014)}]{simonyan2014very}
Simonyan, K.; and Zisserman, A. 2014.
\newblock Very deep convolutional networks for large-scale image recognition.
\newblock \emph{arXiv preprint arXiv:1409.1556} .

\bibitem[{Tolias, Avrithis, and J{\'e}gou(2016)}]{tolias2016image}
Tolias, G.; Avrithis, Y.; and J{\'e}gou, H. 2016.
\newblock Image search with selective match kernels: aggregation across single
  and multiple images.
\newblock \emph{International Journal of Computer Vision} 116(3): 247--261.

\bibitem[{Torii et~al.(2015)Torii, Arandjelovic, Sivic, Okutomi, and
  Pajdla}]{torii201524}
Torii, A.; Arandjelovic, R.; Sivic, J.; Okutomi, M.; and Pajdla, T. 2015.
\newblock 24/7 place recognition by view synthesis.
\newblock In \emph{Proceedings of the IEEE Conference on Computer Vision and
  Pattern Recognition}, 1808--1817.

\bibitem[{Torii et~al.(2013)Torii, Sivic, Pajdla, and
  Okutomi}]{torii2013visual}
Torii, A.; Sivic, J.; Pajdla, T.; and Okutomi, M. 2013.
\newblock Visual place recognition with repetitive structures.
\newblock In \emph{Proceedings of the IEEE conference on computer vision and
  pattern recognition}, 883--890.

\bibitem[{Zhou et~al.(2016)Zhou, Khosla, Lapedriza, Oliva, and
  Torralba}]{zhou2016learning}
Zhou, B.; Khosla, A.; Lapedriza, A.; Oliva, A.; and Torralba, A. 2016.
\newblock Learning deep features for discriminative localization.
\newblock In \emph{Proceedings of the IEEE conference on computer vision and
  pattern recognition}, 2921--2929.

\end{thebibliography}

\end{document}